\documentclass[letterpaper, 10 pt, conference]{ieeeconf}  

\IEEEoverridecommandlockouts                              

\usepackage{cite}
\usepackage{graphics}
\usepackage{epsfig}
\usepackage{amsmath}
\usepackage{booktabs}
\usepackage{threeparttable}
\usepackage{amsmath,amssymb,amsfonts}
\usepackage[ruled,linesnumbered,commentsnumbered,vlined]{algorithm2e}

\usepackage{color}

\definecolor{grey}{rgb}{0.5,0.5,0.5}

\usepackage{caption}
\title{\LARGE \bf
CTS-PLL: A Robust and Anytime Framework for Collaborative Task Sequencing and Multi-Agent Path Finding
}

\author{Junkai Jiang$^{1}$, Yitao Xu$^{1}$, Ruochen Li$^{1}$, Shaobing Xu$^{1}$, and Jianqiang Wang$^{1*}$
}

\begin{document}

\maketitle
\thispagestyle{empty}
\pagestyle{empty}

\begin{abstract}
The Collaborative Task Sequencing and Multi-Agent Path Finding (CTS-MAPF) problem requires agents to accomplish sequences of tasks while avoiding collisions, posing significant challenges due to its combinatorial complexity. This work introduces CTS-PLL, a hierarchical framework that extends the configuration-based CTS-MAPF planning paradigm with two key enhancements: a lock agents detection and release mechanism leveraging a complete planning method for local re-planning, and an anytime refinement procedure based on Large Neighborhood Search (LNS). These additions ensure robustness in dense environments and enable continuous improvement of solution quality. Extensive evaluations across sparse and dense benchmarks demonstrate that CTS-PLL achieves higher success rates and solution quality compared with existing methods, while maintaining competitive runtime efficiency. Real-world robot experiments further demonstrate the feasibility of the approach in practice.

\end{abstract}

\section{Introduction}
\label{Sec:intro}

Multi-Agent Path Finding (MAPF) studies how multiple agents move from given start locations to goal locations without collisions. In many practical scenarios, however, agents must additionally complete a sequence of tasks that may involve cooperative execution. This more specific problem, known as Collaborative Task Sequencing and Multi-Agent Path Finding (CTS-MAPF) \cite{jiang2025ctscbs}, integrates task sequence with motion planning in a unified framework. Formally, CTS-MAPF requires agents to accomplish sequences of tasks while ensuring collision-free paths and eventually reaching their designated goals. By coupling combinatorial task-order decisions with multi-agent path planning, CTS-MAPF greatly enlarges the search space and is therefore significantly harder to solve efficiently and reliably than classical MAPF.




The CTS-MAPF problem captures the essence of many real-world applications, such as robotic fulfillment centers \cite{wurman2008kiva}, coordinated fleets of service robots \cite{salvado2022fleet,dai2024fleet}, and autonomous vehicles operating in structured environment \cite{xu2023distributed}. In these scenarios, agents must complete multiple dependent tasks while avoiding conflicts, which imposes both sequencing and spatial constraints. The key challenge lies not only in handling the exponential growth of the search space, but also in designing solvers that achieve a practical balance between computational efficiency and solution quality. Striking this balance is crucial for applying CTS-MAPF methods to large and complex environments.


\subsection{Related Work} 
\subsubsection{Classical MAPF Solvers}
Research on the MAPF problem has produced a wide variety of solution strategies. Centralized approaches, such as Conflict-Based Search (CBS) \cite{sharon2015cbs}, and its extensions like ICBS \cite{boyarski2015icbs}, ECBS \cite{barer2014suboptimal} and EECBS \cite{li2021eecbs}, can guarantee optimal or bounded-suboptimal solutions. However, their scalability is limited due to the exponential growth of conflicts in dense environments. To address this, decentralized methods such as prioritized planning \cite{erdmann1987prioritized}, rule-based heuristics \cite{sun2014behavior}, and reinforcement learning–based frameworks \cite{sartoretti2019primal,yang2024attention} have been proposed, which improve computational efficiency but lack completeness or optimality guarantees.

In addition, configuration-based solvers have emerged as an efficient alternative. Priority Inheritance with Backtracking (PIBT) \cite{okumura2022priority} allows agents to plan iteratively with lightweight local decisions, achieving remarkable scalability. More recently, LaCAM \cite{okumura2023lacam} was introduced as a search-based configuration method, providing completeness while maintaining efficiency. Nevertheless, LaCAM has yet to be explored in the CTS-MAPF context, where task sequencing constraints impose additional challenges.

\subsubsection{Path Refinement and Anytime Approaches}
Beyond the generation of feasible solutions, improving solution quality has been another active research direction. Large Neighborhood Search (LNS) has been applied to MAPF as an anytime refinement mechanism \cite{li2021anytime}, enabling iterative improvement by re-planning subsets of agents. Variants of LNS have been combined with PIBT or LaCAM to reduce flowtime or makespan in complex environments \cite{jiang2024scaling}, \cite{okumura2024lacamstar}. More generally, anytime algorithms provide a framework to quickly produce feasible solutions and then refine them given more time. While these strategies have shown success in MAPF, their application to CTS-MAPF remains unexplored.

\subsubsection{CTS-MAPF and Related Problems}
The CTS-MAPF problem explicitly integrates collaborative task sequencing with MAPF, making it more challenging than classical MAPF. CBSS \cite{ren2023cbss} has been proposed as a combinatorial extension of CBS, achieving task assignment, scheduling and path planning through the combination of TSP (Traveling Salesman Problem) and CBS, but its scalability remains limited in dense environments. CTS-CBS \cite{jiang2025ctscbs} extends CBS to handle CTS-MAPF, achieving optimal or bounded-suboptimal solutions but also suffering from scalability due to high branching factors. CTS-PIBT \cite{jiang2025ctspibt} introduced a configuration-based solver for CTS-MAPF, demonstrating superior efficiency and success rates in large-scale instances. However, as a PIBT-based method, it remains incomplete and offers limited support for solution refinement.

Other related problems share similarities with CTS-MAPF but differ in formulation. Multi-goal MAPF (MG-MAPF) \cite{surynek2021mgmapf} requires agents to visit multiple goals but does not distinguish tasks from goals. Multi-agent pickup and delivery (MAPD) \cite{camisa2022mapd} involves task assignment and ordering but restricts tasks to paired pickup–delivery operations. Integrated task sequencing and path planning problems \cite{xu2022multi,honig2018conflict} consider task assignment and scheduling jointly, but typically lack the generality needed for CTS-MAPF. These distinctions highlight the uniqueness of CTS-MAPF and the need for dedicated methods.

\subsection{Motivations and Contributions} 

Although recent progress has been made, CTS-MAPF remains challenging. Existing solvers based on CBS ensure high-quality solutions but scale poorly in complex environments. Configuration-based approaches such as CTS-PIBT offer high efficiency but are incomplete. Meanwhile, path refinement techniques such as LNS and anytime search have demonstrated strong potential in MAPF but are rarely incorporated into CTS-MAPF frameworks.

To address these limitations, we propose a new CTS-MAPF solver that augments CTS-PIBT with two key components. First, a lock agents detection module identifies stagnated agents and employs MAPF solvers with completeness for local re-planning, maintaining efficiency while improving robustness. Second, we extend the anytime mechanism by incorporating an LNS-based refinement procedure after each planning cycle, allowing the algorithm to continuously improve solution quality over time.

The contributions of this work are summarized as follows:

\begin{enumerate}
    
    \item We develop a novel CTS-MAPF solver that enhancing the success rate of configuration-based method through the lock agents detection and release mechanism.
    
    
    \item We introduce an enhanced anytime mechanism that integrates an LNS-based refinement step, enabling progressive improvements in solution quality.
    
    \item We conduct extensive evaluations in various environments to demonstrate the superior performance of our algorithm, and perform practical robot tests to validate its effectiveness in real-world scenarios.
\end{enumerate}

\section{Preliminaries}
\label{Sec:pre}

\subsection{Problem Definition}
\label{Sec:sub-pro-def}

The CTS-MAPF problem has been formally defined in \cite{jiang2025ctspibt}. 
Here, we briefly restate the essential elements and notations to keep this paper self-contained.  

The environment is modeled as an undirected graph $G=(V,E)$, where $V$ is the set of vertices and $E \subseteq V \times V$ denotes the edges. 
Time is discretized into steps $t=0,1,\ldots$. 
A set of agents $\mathcal{A}=\{a_1,\ldots,a_m\}$ is deployed, where each agent $a_j$ has an initial position $s_j \in V$ and a terminal goal $g_j \in V$. 
The path of agent $a_j$ is denoted by $\pi_j$, where $\pi_j(t)$ represents the vertex occupied by agent $a_j$ at $t$.
At each time step, an agent may either remain at its current vertex or move to a neighboring one. 
Two types of conflicts are prohibited: (i) \textbf{vertex conflict}, i.e., $\pi_j(t)=\pi_k(t)$ for some $j \neq k$; and (ii) \textbf{edge conflict}, i.e., $(\pi_j(t),\pi_j(t+1))=(\pi_k(t+1),\pi_k(t))$ for some $j \neq k$.  

CTS-MAPF involves a set of tasks $V_t=\{v_{t1},\ldots,v_{tn}\}$. 
Each task $v_{ti}$ corresponds to a specific vertex that must be visited by a designated subset of agents $f(v_{ti})\subseteq \mathcal{A}$. 
Each agent is required to complete all of its targets, which include visiting every task vertex assigned to it and reaching its terminal goal.
Thus, solving CTS-MAPF requires determining: (1) the task sequence of each agent, and (2) joint path $\pi_j$ such that $\pi_j(0)=s_j$, all assigned tasks are completed sequentially, and terminates at $g_j$.  

A feasible solution of CTS-MAPF is a joint path $\Pi=\{\pi_1,\ldots,\pi_m\}$ that satisfies three conditions: (i) all tasks $v_{ti}$ are completed while each corresponding agent has visited them at least once; (ii) agents do not experience either vertex or edge conflicts during execution; and (iii) every agent $a_j$ starts at $s_j$ and ends at $g_j$.  

The quality of a solution is commonly evaluated by cost functions such as the \textit{makespan} $C_{\max}=\max_{j=1}^m T_j$, where $T_j$ is the arrival time of agent $a_j$ at $g_j$, or the \textit{flowtime} $C_{\Sigma}=\sum_{j=1}^m T_j$. In this work, we adopt \textit{flowtime} as the primary metric for evaluating solution quality.

\subsection{Review of CTS-PIBT}
CTS-PIBT was proposed as a baseline framework for the CTS-MAPF problem, combining task sequencing with a lightweight MAPF solver. It follows a hierarchical structure consisting of two main components: generating task sequences for agents, and conducting low-level path finding following each task sequence with an extended version of PIBT. This design leverages the efficiency of PIBT-based planning while maintaining the ability to explore multiple task sequences, enabling the algorithm to achieve high success rates and scalability in large-scale scenarios.

At the path-planning level, CTS-PIBT uses Extended-PIBT to generate collision-free joint paths. Extended-PIBT follows the stepwise PIBT paradigm: at each discrete timestep $t$, it maintains a configuration $Q_t=(v_1,\ldots,v_m)$ of all agent positions, and each agent selects its next vertex from adjacent vertices (or waits in place). Agents are processed in priority order with priority inheritance: if an agent requests a vertex occupied by another agent, the occupying agent is planned first via recursive backtracking, which helps resolve tight local conflicts. When no valid move exists, the agent waits. The procedure is summarized in Algorithm~\ref{alg:ExtendedPIBTStep}. With task-aware priority design and modified termination conditions, Extended-PIBT extends PIBT to meet CTS-MAPF requirements and, together with the task-sequencing layer, forms a lightweight framework that integrates task sequencing with online collision avoidance.

\begin{algorithm}[t]
    \caption{Extended-PIBT Step}
    \label{alg:ExtendedPIBTStep}
    \KwIn{Current config $Q_f$, agents $\mathcal{A}$, priorities $P$}
    \KwOut{Next config $Q_t$}
    
    $Q_t \gets \text{each element initialized as} \bot$ \;
    
    Sort agents in descending order of $P$ \;

    \For{$a_i \in \mathcal{A}$}{
        \lIf{$Q_t[i] = \bot$}{\texttt{funcExPIBT}$(Q_f, Q_t, i)$}
    }
    
    \Return $Q_t$ \;
    
    \SetKwFunction{funcExPIBT}{funcExPIBT}
    \SetKwProg{Fn}{function}{:}{}
    \Fn{\funcExPIBT{$Q_f$, $Q_t$, $i$}}{
        $C \gets \text{neigh}(Q_f[i]) \cup \left\{ Q_f[i] \right\}$  \;
        $v_{\text{next}} \gets$ next unvisited task location or goal of $a_i$ \;
        Sort $C$ in ascending order of distance to $v_{\text{next}}$ \;
        
        \For{$v \in C$}{
            \lIf{\texttt{hasConflict}($Q_t, i, v$)}{\textbf{continue}}
            $Q_t[i] \gets v$ \;

            \If{$\exists a_j \neq a_i$ s.t. $Q_f[j] = v \land Q_t[j] = \bot$}{
                \lIf{\funcExPIBT{$Q_f, Q_t, j$} = invalid}{\textbf{continue}}
            }
            \Return valid
        }
        $Q_t[i] \gets Q_f[i]$;
        \Return invalid
    }
\end{algorithm}

\begin{figure}[htbp]
	\centering
	\includegraphics[width=0.95\linewidth]{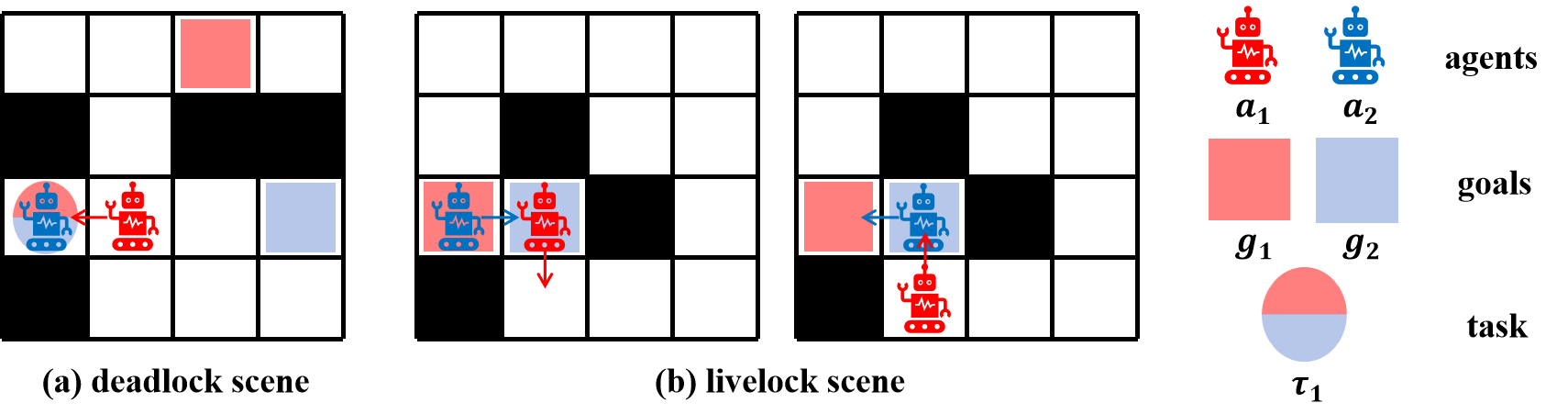}
	\caption{Examples of deadlock and livelock cases in CTS-PIBT. (a) Deadlock: agent $a_1$ attempts to execute task $v_{t1}$, but agent $a_2$ has no alternative position to yield, blocking the move. (b) Livelock: agents $a_1$ and $a_2$ repeatedly oscillate around each other's goals, preventing convergence.}
	\label{fig:lock_scenes}
\end{figure}

By design, CTS-PIBT is a lightweight centralized framework that scales well to large-scale agents and can generate feasible solutions in real time. Its step-by-step, configuration-based procedure enables agents to coordinate locally at each timestep, which often resolves conflicts efficiently in practice. Nevertheless, PIBT is neither optimal nor complete; consequently, CTS-PIBT may fail to recover once local conflicts escalate into deadlocks or livelocks. For example, in Fig.~\ref{fig:lock_scenes}(a), agent $a_1$ is assigned a higher priority because it is farther from its final goal and thus attempts to move left to execute task $v_{t1}$. However, agent $a_2$ has no feasible position to yield; it returns \textit{invalid} for this request and stays in place. Consequently, $a_1$ falls back to its second-best action and also waits (since staying is closer to $v_{t1}$ than moving up, down, or right), leaving both agents stationary and resulting in a deadlock. In Fig.~\ref{fig:lock_scenes}(b), agents $a_1$ and $a_2$ temporarily occupy each other's goal locations. If $a_2$ has higher priority, it reaches its goal while forcing $a_1$ downward. In the next step, priorities swap, causing $a_1$ to move upward and push $a_2$ leftward. This priority alternation repeats, leading to persistent oscillation and a livelock. Moreover, the greedy task scheduling strategy in CTS-PIBT has only a limited effect on improving global solution quality. To address these issues, this work introduces two key enhancements: (i) a lock-agent detection module that identifies agents trapped in deadlock/livelock patterns and triggers a localized re-planning procedure based on LaCAM, thereby improving robustness and completeness; and (ii) a new anytime framework that incorporates an LNS-based refinement after each planning phase, enabling continuous improvement of solution quality over time.

\section{Method}
\label{Sec:method}

In this section, we first present the overall process of our CTS-PLL (Enhanced CTS-PIBT with LaCAM and LNS) algorithm. We then provide detailed descriptions of its two core components: the lock detection and release module, and the LNS-based quality improvement module.

\subsection{CTS-PLL Algorithm}
CTS-PLL follows the same hierarchical structure as CTS-PIBT, consisting of joint task sequencing and low-level path finding. Building upon this framework, CTS-PLL further introduces dedicated mechanisms to improve robustness and continuously enhance solution quality.

\begin{figure}[htbp]
	\centering
	\includegraphics[width=0.95\linewidth]{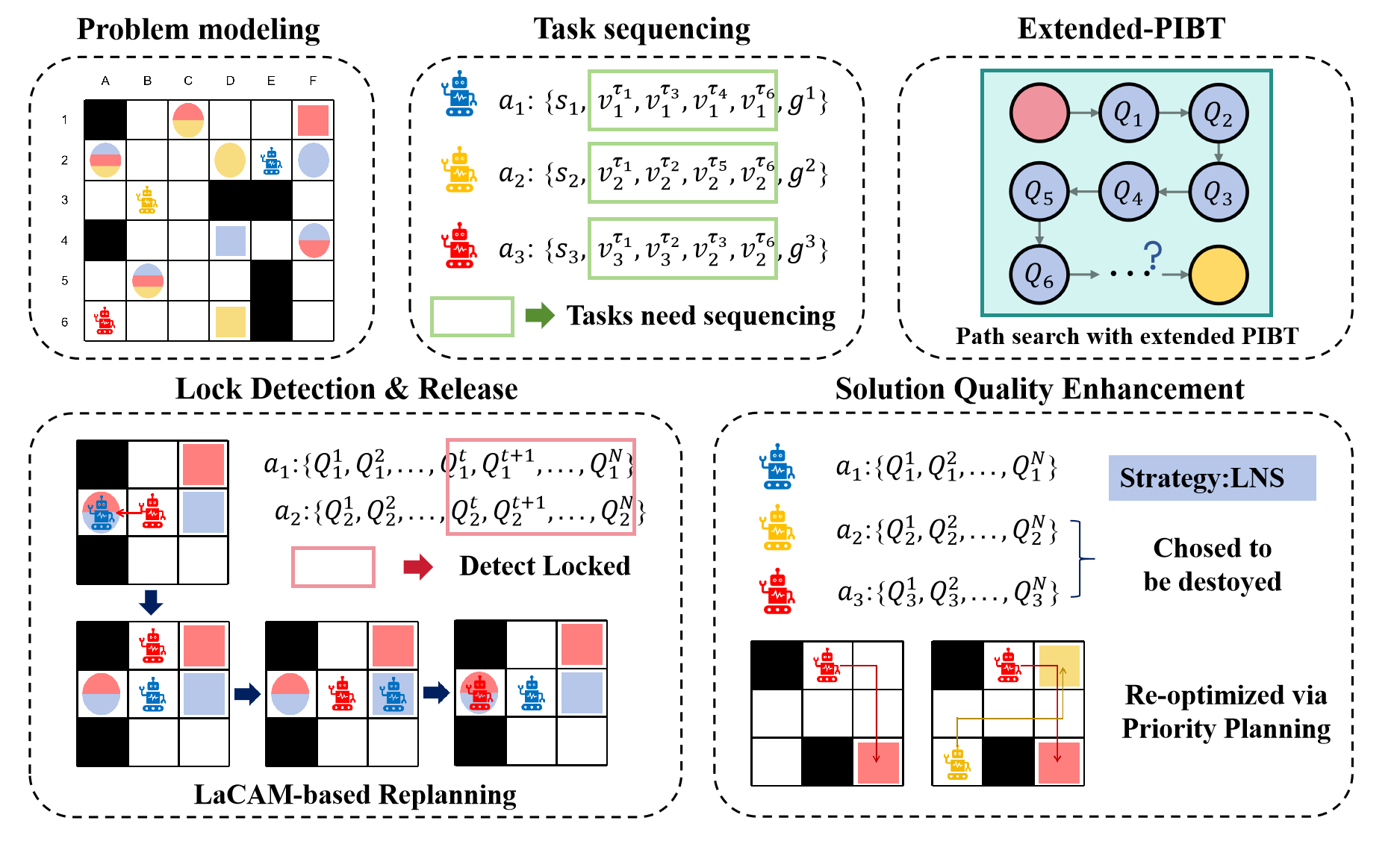}
	\caption{The main process of CTS-PLL, which has four key components: joint task sequencing, path finding using Extended-PIBT, lock detection and release module and solution quality enhancement via LNS.}
	\label{fig:CTSPLL}
\end{figure}

As shown in Fig.~\ref{fig:CTSPLL}, CTS-PLL first generates the top-$K$ joint task sequences using the bi-level procedure proposed in \cite{jiang2025ctspibt}. For each candidate sequence, it runs Extended-PIBT augmented with a lock-detection module. If Extended-PIBT fails to produce a valid solution within a predefined step limit, lock detection is triggered to identify stagnated agents. The algorithm then invokes LaCAM for localized MAPF problem to escape from the lock, which will be introduced in Sec~\ref{sec:lock-agents-detectionRelease}

In addition, CTS-PLL incorporates an anytime refinement procedure to improve path quality. For each task sequence under consideration, the planner first generates collision-free joint paths and then applies LNS with a fixed budget (either a preset number of iterations or a time limit) to refine the solution. It subsequently moves on to the next task sequence and repeats the same refinement process. In each LNS iteration, a subset of agent paths is destroyed and re-optimized, enabling the solution to be gradually improved (e.g., toward lower flowtime). The refinement terminates once the budget is exhausted. As a result, CTS-PLL can quickly output feasible solutions and, when additional time is available, progressively improve them.

Formally, the overall workflow of CTS-PLL is summarized in Algorithm~\ref{alg:cts-pll}. The algorithm first initializes the sequence index $i=1$ and sets the current best solution $\Pi^*$ to \emph{None}. Then, within the time limit $T_{\max}$, it iteratively explores candidate joint task sequences. At each iteration, CTS-PLL queries \emph{KBestJointSequencing} with $K=i$ to obtain the current best task sequence $\mathcal{T}_i^*$, and computes a corresponding joint plan $\Pi_i$ by running \emph{ExtendedPIBT}. If the returned plan $\Pi_i$ is infeasible, the planner repeatedly triggers the \emph{LockDetect\&Release} procedure to update $\Pi_i$ and resumes \emph{ExtendedPIBT} until a feasible solution is obtained. Once feasibility is achieved, CTS-PLL applies an LNS-based refinement to $\Pi_i$, producing an improved plan $\Pi'_i$. The $\Pi^*$ is updated whenever $\Pi'_i$ has a lower cost than the current best (or when $\Pi^*$ is \emph{None}). The process continues with subsequent task sequences until the time budget is exhausted, and the algorithm finally returns the lowest-cost solution $\Pi^*$ found within $T_{\max}$.

\begin{algorithm}[t]
\caption{CTS-PLL Algorithm}
\label{alg:cts-pll}
\KwIn{Graph $G$, agents $\mathcal{A}$, starts $\mathcal{S}$, goals $\mathcal{G}$, tasks $V_t$, time limit $T_{\max}$}
\KwOut{Joint Paths $\Pi^*$}
$i=1$; best solution $\Pi^* \gets$ None\;
\While{t $< T_{\max}$}{
    $\mathcal{T}_i^* \gets KBestJointSequencing(K=i)$\;
    $\Pi_i \gets ExtendedPIBT(\mathcal{T}_i^*)$\;
    \While{\rm{$\Pi_i$ is not feasible}}{
        updated $\Pi_i \gets LockDetect\&Release$ and resume $ExtendedPIBT$\;
    }
    $\Pi'_i \gets$ refine $\Pi_i$ by LNS\;
    \If{\rm{cost$(\Pi'_i) <$ cost$(\Pi^*)$ or $\Pi^*$ is None}}{
        update $\Pi^* \gets \Pi'_i$\;
    }
    $i=i+1$\;
}
\Return $\Pi^*$ \;
\end{algorithm}

\subsection{Lock Agents Detection and Release} \label{sec:lock-agents-detectionRelease}

As discussed above, the purpose of the lock agents detection and release stage is to help agents escape from lock states. Therefore, as shown in Line 6 of Algorithm~\ref{alg:cts-pll}, this stage outputs an updated joint path $\Pi'$. Note that $\Pi'$ is not a complete CTS-MAPF solution; it only resolves the lock by releasing the stagnated agents. The planner must subsequently resume Extended-PIBT to obtain a full joint path for the CTS-MAPF problem. The procedure of lock detection and release is shown in Algorithm~\ref{alg:lock-detect-release}.

\begin{algorithm}[t]
    \caption{LockDetect\&Release}
    \label{alg:lock-detect-release}
    \KwIn{Infeasible path $\Pi$, goals $\mathcal{G}$, tasks $V_t$}
    \KwOut{Updated incomplete path $\Pi'$}
    $\mathcal{A}_{\text{lock}}\gets DetectLock(\Pi,\mathcal{G})$\;
    $\hat{t}\gets FindLockTime(\Pi,\mathcal{A}_{\text{lock}},\mathcal{G},V_t)$\;
    $\widetilde{Q}_{\hat{t}} \gets$ configuration of $\mathcal{A}_{\text{lock}}$ at $\hat{t}$\;
    \For{$a_i \in \mathcal{A}_{\text{lock}}$}{
        $\widetilde{Q}_{\text{target}}[i] \gets$ define the target vertex of $a_i$\;
    }
    $\widetilde{\Pi} \gets LaCAM(\widetilde{Q}_{\hat{t}},\widetilde{Q}_{\text{target}},\mathcal{A}_{\text{lock}})$\;
    $\Pi' \gets$ use $\widetilde{\Pi}$ to update $\Pi$\;

    \Return $\Pi'$\;
\end{algorithm}

\subsubsection{Lock Agents Detection and Time Localization}
The lock detection mechanism is triggered whenever Extended-PIBT returns an infeasible joint path plan. It then identifies the lock-agent set $\mathcal{A}_{\text{lock}}$ from this plan as follows (Line 1 in Algorithm~\ref{alg:lock-detect-release}): (i) all agents that have not yet completed assigned tasks and reached their final goals; and (ii) agents that have already reached their goals but occupy locations that block the motion of unfinished agents. This definition captures both active agents that are still executing tasks and passive agents that inadvertently cause stagnation.

After identifying the lock-agent set $\mathcal{A}_{\text{lock}}$, we further determine the time at which the system enters the lock state (Line 2 in Algorithm~\ref{alg:lock-detect-release}). Specifically, for each agent $a_i \in \mathcal{A}_{\text{lock}}$, we compute $t_i^{\text{last}}$, the timestep when $a_i$ completes its most recently finished task in the (infeasible) joint path returned by Extended-PIBT. The lock time is then defined as: $\hat{t} = \min_{a_i \in \mathcal{A}_{\text{lock}}} t_i^{\text{last}}$. By taking the minimum, we locate the earliest moment at which progress among the lock agents may start to stall, so that a recovery procedure can be applied as early as possible. We denote the configuration of the lock agents at time $\hat{t}$ as $\widetilde{Q}_{\hat{t}}$ (Line 3 in Algorithm~\ref{alg:lock-detect-release}), where the tilde indicates that the configuration only includes agents in $\mathcal{A}_{\text{lock}}$. The recovery module is then applied to $\widetilde{Q}_{\hat{t}}$ to resolve the lock.

\subsubsection{Lock Release via MAPF Solver with Completeness}

After determining the lock-agent set $\mathcal{A}_{\text{lock}}$ and the lock time $\hat{t}$, we release the stagnation by solving a local MAPF instance over the lock agents only. In this stage, agents outside $\mathcal{A}_{\text{lock}}$ are frozen at their current vertices and treated as static obstacles, so the re-planning is confined to the congested region while leaving the remaining agents unaffected. Any complete MAPF solver can be used to compute a collision-free local path. In this work we adopt LaCAM.

Lines 4–5 of Algorithm~\ref{alg:lock-detect-release} construct the local MAPF instance. The initial configuration is $\widetilde{Q}_{\hat{t}}$, and the key is to define the target configuration $\widetilde{Q}_{\text{target}}$. A straightforward choice is to set $\widetilde{Q}_{\text{target}}[i]$ as the next task vertex in agent $a_i$'s task sequence (or its final goal if all tasks have been completed), which helps agents escape while preserving the prescribed task order. However, we observe that this target configuration definition often leads to high computational overhead in practice: LaCAM may require a long time to succeed, likely because its configuration generator is also PIBT-based and must enumerate many candidate configurations when the system is already in a locked state.

Therefore, in practical runs we instead construct $\widetilde{Q}_{\text{target}}$ by randomly exchanging the positions of agents in $\mathcal{A}_{\text{lock}}$ (i.e., applying a random permutation over $\widetilde{Q}_{\hat{t}}$). This perturbation breaks common symmetries behind deadlocks/livelocks and redistributes agents away from heavily congested vertices, thereby creating new opportunities for subsequent planning. While this target selection does not provide a formal completeness guarantee of lock release, it is empirically effective at resolving the majority of stagnation cases (See Section~\ref{Sec:exp}). Once LaCAM returns a local plan $\widetilde{\Pi}$, we splice $\widetilde{\Pi}$ into the original infeasible plan $\Pi$ to obtain an updated (here still incomplete) plan $\Pi'$, recompute agent priorities based on the new configuration, and resume Extended-PIBT to search for a feasible CTS-MAPF solution.

\subsection{Solution Quality Enhancement via LNS}

CTS-PIBT provides an anytime mechanism by enumerating alternative joint task sequences and invoking Extended-PIBT to obtain a path plan for each sequence. However, this form of anytime improvement is \emph{sequence-only}: it relies solely on switching to a different task sequence and does not actively optimize the path quality under a fixed sequence. Therefore, in CTS-PLL, after Extended-PIBT (with lock detection and release) returns a feasible joint plan for a considered sequence, we further apply Large Neighborhood Search to refine the plan cost.

Given a feasible joint plan $\Pi$, LNS iteratively performs a destroy-repair cycle under a fixed budget (either a preset number of iterations or a time limit). In the destroy phase, we select a subset of agents and remove their paths from $\Pi$, while keeping the remaining paths fixed. Here, we employ three complementary neighborhood-selection strategies:
(i) \textbf{Random}: uniformly sample agents to maintain exploration diversity;
(ii) \textbf{Intersection-based}: prioritize agents whose paths traverse congested (high-conflict) vertices;
(iii) \textbf{Random-walk}: start from a delayed agent and follow conflicts along its trajectory to construct a correlated neighborhood.

To adaptively allocate search effort among these strategies, we maintain a non-negative weight $w_s$ for each strategy $s$. After applying strategy $s$ once, we compute its \emph{local cost saving}
\[
\Delta_s \;=\; \sum_{\pi \in \Pi_s^-} C(\pi)\;-\;\sum_{\pi \in \Pi_s^+} C(\pi),
\]
where $\Pi_s^-$ denotes the set of paths removed (before re-planning) and $\Pi_s^+$ denotes the repaired paths produced for the same agents, and $C(\cdot)$ is the path cost. A positive $\Delta_s$ indicates that the repair reduces the cost on the affected subset. We then update the weight using an exponential moving average:
\[
w_s \leftarrow (1-\gamma) w_s + \gamma \max\{\Delta_s,0\},
\]
where $\gamma \in [0,1]$ controls the update rate. This rule rewards strategies that yield consistent improvements and gradually downplays those that rarely help.

In the repair phase, the removed agents are re-planned sequentially using prioritized planning (PP), while the preserved paths are treated as spatiotemporal obstacles to avoid both vertex and edge conflicts. If the repaired plan $\Pi'$ is feasible and improves the current best plan, we accept it; otherwise, we keep the current one. Since LNS can be terminated at any time budget, CTS-PLL inherits an anytime property: it can output a feasible solution quickly and progressively reduce the flowtime as refinement continues, which is particularly effective in congested settings.


\section{Experiments and Validation}
\label{Sec:exp}

We conducted extensive experiments to evaluate the effectiveness of CTS-PLL in both simulation and real-world settings. The evaluation was performed under two representative types of grid-based maps: sparse scenarios with relatively low agent density, and dense scenarios with frequent interactions and congestion. In each case, we compared CTS-PLL against baseline methods to assess its performance in terms of success rate, running time, and flowtime. Beyond simulation, we further validated the practicality of CTS-PLL on a set of physical mobile robots, demonstrating its applicability in real-world environments.

\subsection{Settings, Benchmark and Baselines}

All experiments were conducted on a computer equipped with an Intel Core i7-12700KF CPU and 32GB of RAM, running on a Linux system. The implementation of CTS-PLL was developed in C++, and all baseline methods were executed under the same environment to ensure fairness.

\begin{figure}[htbp]
	\centering
	\includegraphics[width=0.95\linewidth]{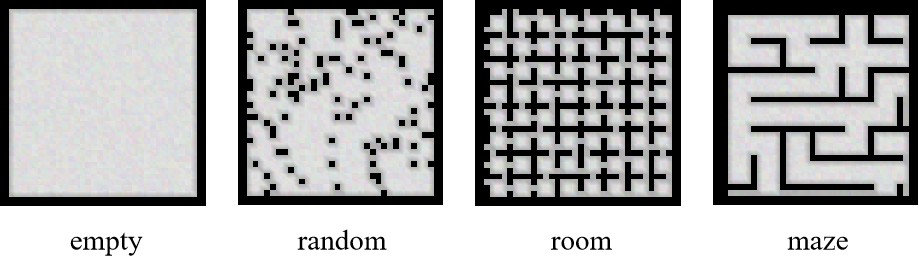}
	\caption{Benchmark maps adopted for evaluation, including four representative scenarios: empty, random, room, and maze. These maps are used to construct sparse and dense experimental settings.}
	\label{fig:benchmark}
\end{figure}

For the benchmark maps, we used four standard grid scenarios from a CTS-MAPF dataset proposed in \cite{jiang2025ctscbs}: \emph{random}, \emph{room}, \emph{maze}, and \emph{empty}, as shown in Fig.~\ref{fig:benchmark}. Sparse settings were created on all four maps, while dense settings were generated only on the \emph{maze} map, since it is the most challenging scenario and most prone to agent conflicts. For every combination of agent and task numbers (N, M) under each setting, we generated 50 random instances with different start, goal, and task locations. The detailed experimental configurations, including the ranges of agent and task numbers for each setting, are summarized in Table~\ref{tab:settings}.

To clearly distinguish the contributions of different components in our approach, we evaluate three variants of CTS-PLL in the experiments:

\textbf{CTS-PLL-v1}: Uses only the best task sequence and incorporates only the lock detection and release module on top of CTS-PIBT. 

\textbf{CTS-PLL-v2}: An early-stop version that includes the lock detection and release module and searches through task sequences until a feasible solution is found, but does not include LNS optimization.

\textbf{CTS-PLL-v3}: The full anytime version that includes both lock detection and release and LNS modules. It continues the search and refinement until a preset time threshold, allowing continuous improvement of solution quality over time.

In terms of baselines, for the sparse setting we compared CTS-PLL against three representative methods: CTS-CBS \cite{jiang2025ctscbs}, CBSS \cite{ren2023cbss}, and CTS-PIBT \cite{jiang2025ctspibt}. In the dense setting, we only compared against CTS-PIBT, as the success rates of CTS-CBS and CBSS were too low to yield meaningful results. For runtime limits, in the sparse setting both CTS-PLL-v3 and CTS-PIBT were given a limit of 60s, while CBSS and CTS-CBS were allowed up to 180s. In the dense setting, both CTS-PLL-v3 and CTS-PIBT-anytime were allocated 180s.

\begin{table}[t]
\centering
\caption{Experimental settings for sparse and dense scenarios.}
\label{tab:settings}
\resizebox{\columnwidth}{!}{%
\begin{tabular}{lcccc}
\toprule
\textbf{Scenario type}  & \textbf{Maps} & \textbf{Agents $N$} & \textbf{Tasks $M$} & \textbf{Instances} \\
\midrule
Sparse & four maps & $5,\,10,\,20$ & $10,\,20,\,30,\,40,\,50$ & 50 \\
Dense  & maze & $50$ & $60,\,90,\,120,\,150,\,180$ & 50 \\
\bottomrule
\end{tabular}%
}
\end{table}

\begin{figure*}[bthp]
    \begin{center}    
         \includegraphics[trim=0 0 0 0, clip, width=0.92\linewidth]{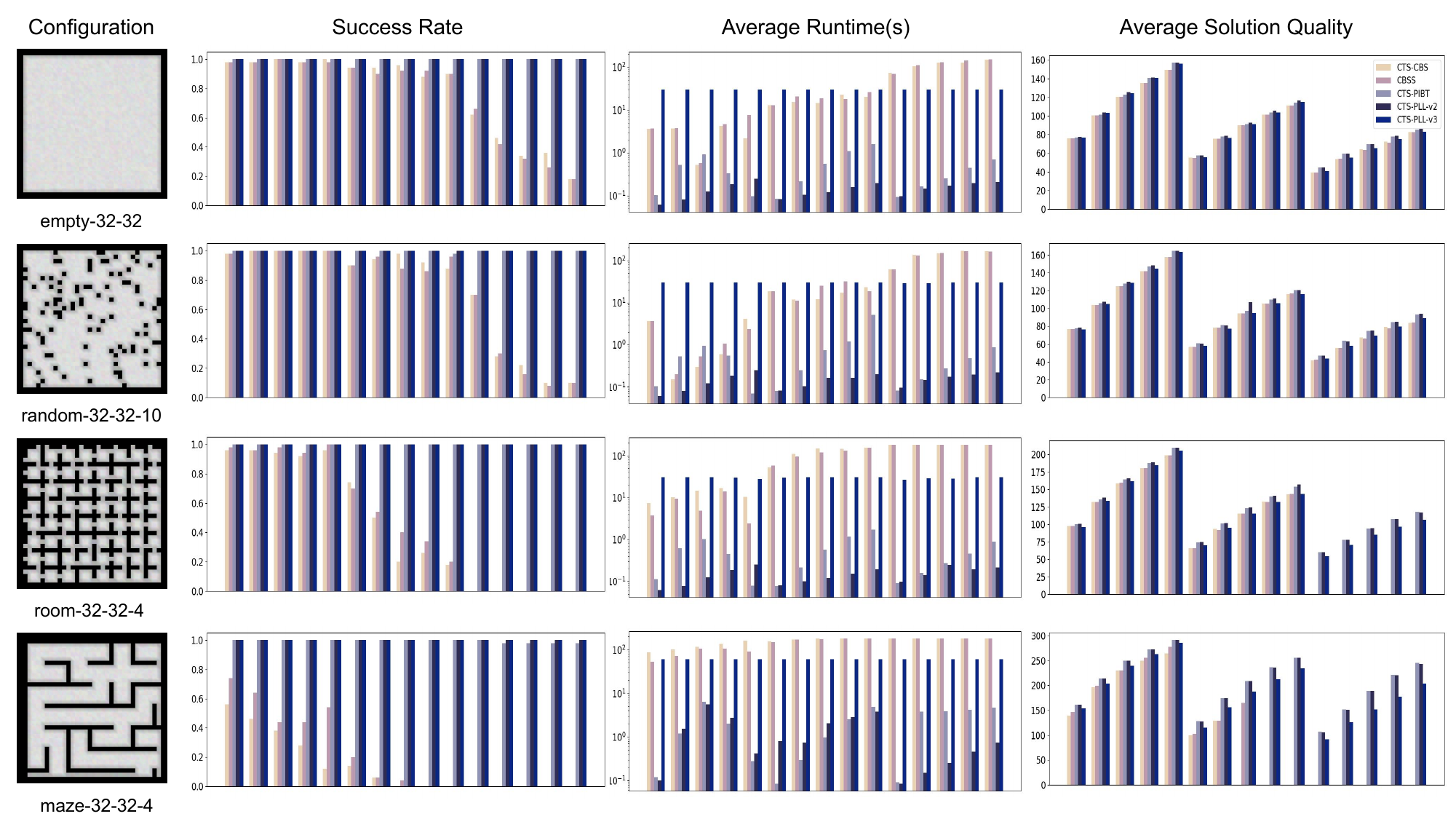}
    \end{center}
    \caption{Performance of CTS-PLL and baseline methods under different difficulty levels (map types, agent numbers, and task numbers). For clarity, we omit the x-axis labels in each subfigure since they share the same contents. The x-axis represents the pairs of (N,M) (agent number, task number), ranging from (5,10) to (5,50), (10,10) to (10,50), and (20,10) to (20,50). Here, CTS-PLL-v2 denotes the variant without anytime optimization, while CTS-PLL-v3 refers to the full method with anytime mechanism and LNS refinement.}
    \vspace{-3mm}
    \label{fig:comparison}
\end{figure*}

\vspace{-1mm}
\subsection{Comparative Analysis in Sparse Scenes}
We first evaluate CTS-PLL in the sparse settings. The comparative results are summarized in Fig.~\ref{fig:comparison}, where performance is reported in terms of success rate, average runtime, and average solution quality (flowtime). 

In terms of success rate, both two versions of CTS-PLL demonstrate clear superiority, successfully solving all tested cases across maps and parameter scales in sparse scenes. Even in instances where CTS-PIBT failed due to unresolved deadlock or livelock, CTS-PLL-v2 and CTS-PLL-v3 consistently provided feasible solutions, confirming the effectiveness of the lock detection and recovery module. By contrast, search-based methods such as CTS-CBS and CBSS, despite being given a longer time limit of 180 s, still exhibited significantly lower success rates in more constrained maps.

Regarding computational efficiency, CTS-PLL-v2 is able to complete all cases within a very short time, maintaining efficiency on par with CTS-PIBT while avoiding the exponential growth in runtime observed in search-based approaches. Moreover, the CTS-PLL-v3 was executed with a fixed 60 s budget, during which it continuously refined solutions to improve path quality, demonstrating both fast feasibility and the ability to enhance solution quality given additional time.

As for solution quality, CTS-PLL-v2 already matches or slightly improves upon CTS-PIBT, while the anytime-enhanced version consistently achieves lower costs across all benchmark maps. The improvement is most evident in structured environments such as \emph{room} and \emph{maze}, where congestion is more likely to accumulate. 

Overall, the results in sparse scenarios highlight the strengths of CTS-PLL: it preserves the scalability and efficiency of Extended-PIBT, guarantees 100\% success rate by overcoming deadlock and livelock failures, and, with the anytime LNS refinement, further improves solution quality under identical runtime constraints.

\subsection{Denser Configuration Experiments}

To more clearly evaluate the contribution of each enhancement module in CTS-PLL, we conduct ablation studies in denser configurations, where agent-agent interactions are much more frequent. Experiments are performed on the most complex scenario \emph{maze} map with 50 agents and task numbers varying from 60 to 180. We only compare CTS-PLL variants against CTS-PIBT, as other baselines fail to solve most instances in these dense settings.

\begin{table}[htpb]
    \caption{Success rate results of denser configurations on maze map.} \label{tab:ablation}
    \label{tab:dense_success}
    \vspace{-2mm}
    \centering
    \resizebox{\linewidth}{!}{
    \begin{tabular}{c ccccc}
    \toprule
    Method    & \multicolumn{5}{c}{Success Rate $\uparrow$} \\
    \cmidrule(lr){2-6}
    \# Tasks  & 60 & 90 & 120 & 150 & 180 \\
    \midrule
    CTS-PIBT & 76\% & 68\% & 66\% & 50\% & 38\% \\
    CTS-PLL-v1 & 98\% & 92\% & 96\% & 92\% & 72\% \\
    \midrule
    Improvement Rate & 28.95\% & 35.29\% & 45.45\% & 84.00\% & 89.47\% \\
    \bottomrule
    \end{tabular}
    }
    \begin{tablenotes}[para,flushleft,small]
    \raggedright
    \scriptsize 
    \item Note: Success rate comparison between CTS-PIBT and CTS-PLL-v1 on the \emph{maze} map with 50 agents. CTS-PLL-v1 incorporates only the lock detection and release module, demonstrating that this mechanism effectively resolves deadlock and livelock situations, leading to significantly higher success rates as task density increases.
    \end{tablenotes}
    \vspace{2mm}
\end{table}

First we compare CTS-PLL-v1 with CTS-PIBT in terms of success rate. They both only search feasible solution under the best task sequence, with CTS-PLL-v1 augmenting CTS-PIBT solely through the lock detection and release module. As shown in Table~\ref{tab:dense_success}, CTS-PLL-v1 consistently achieves higher success rates across all task counts, with improvements ranging from 28.95\% to 89.47\%. This confirms that the proposed lock detection and LaCAM-based recovery effectively resolve deadlocks and livelocks that frequently occur in dense settings, thereby substantially enhancing the robustness of the configuration-based planner.

\begin{table}[htpb]
    \caption{Solution quality results of denser configurations on maze map.} \label{tab:ablation}
    \label{tab:dense_quality}
    \vspace{-2mm}
    \centering
    \resizebox{\linewidth}{!}{
    \begin{tabular}{c ccccc}
    \toprule
    Method    & \multicolumn{5}{c}{Solution Quality (Flowtime) $\downarrow$} \\
    \cmidrule(lr){2-6}
    \# Tasks  & 60 & 90 & 120 & 150 & 180 \\
    \midrule
    CTS-PIBT-anytime & 196.35 & 253.21 & 316.93 & 389.62 & 457.04 \\
    CTS-PLL-v3 & 162.00 & 221.89 & 301.85 & 357.99 & 451.34 \\
    \midrule
    Improvement Rate & 17.49\% & 12.37\% & 4.76\% & 8.12\% & 1.25\% \\
    \bottomrule
    \end{tabular}
    }
    \begin{tablenotes}[para,flushleft,small]
    \raggedright
    \scriptsize 
    \item Note: We compare CTS-PIBT-anytime and CTS-PLL-v3 in terms of average solution quality (flowtime). Results show that CTS-PLL-v3 consistently provides shorter flowtime solutions across all tested configurations.
    \end{tablenotes}
    \vspace{0mm}
\end{table}

We then compare the solution quality (flowtime) between CTS-PIBT-anytime and CTS-PLL-v3 in Table~\ref{tab:dense_quality}. Both algorithms will search for feasible solutions throughout candidate task sequences until the time limit, while CTS-PLL-v3 integrates both lock recovery and LNS refinement. The results demonstrate that CTS-PLL-v3 yields shorter flowtimes, with improvements between 1.25\% and 17.49\%. This indicates that the LNS module effectively optimizes the paths after a feasible plan is obtained, leading to better overall efficiency.

Together, these ablation results demonstrate the distinct yet synergistic contributions of the two proposed modules: the lock detect and release module ensures high success rates in challenging dense scenarios, while the anytime LNS refinement progressively improves the solution quality. The combined approach in CTS-PLL provides comprehensive improvements in both feasibility and efficiency for dense CTS-MAPF environments.

\subsection{Practical Robot Tests}
In this section, we conducted physical robot experiments to evaluate the CTS-PLL algorithm. The experimental setup employed toio robots, which maintained Bluetooth communication with a central controller and navigated on a designated mat by following continuous coordinate commands.

\begin{figure*}[ht]
    \begin{center}    
         \includegraphics[trim=0 0 0 0, clip, width=0.92\linewidth]{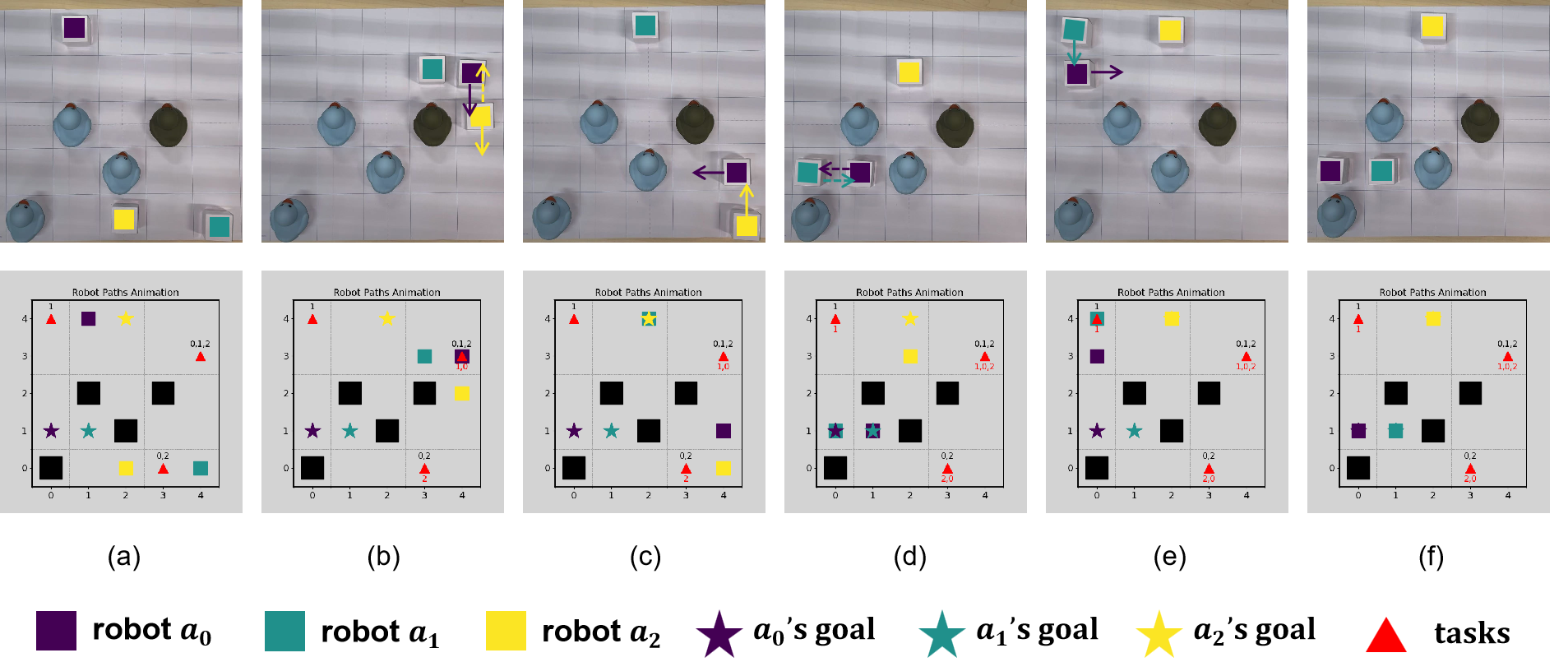}
    \end{center}
    \caption{Physical robot experiment using toio robots to validate CTS-PLL. Robots are shown as purple, cyan, and yellow squares. The corresponding colored stars indicate their goal locations, while red triangles represent task points. Black numbers above each task denote the robots assigned to it, and red numbers below show the robots that have already completed it. The top subfigure illustrates the real-world robot experiment, while the bottom subfigure shows the corresponding simulation process. The figure depicts the trajectories of the robots during the experiment.}
    \label{fig:real-robot}
    \vspace{-3mm}
\end{figure*}

Fig.~\ref{fig:real-robot} presents snapshots from one representative run of the physical robot experiments. Fig.~\ref{fig:real-robot}(a) shows the initial configuration of the robots, while Fig.~\ref{fig:real-robot}(f) depicts the final state after all tasks are completed. During the early stages (Fig.~\ref{fig:real-robot}(b) and \ref{fig:real-robot}(c)), the robots follow the planning of Extended-PIBT. Specifically, agent $a^0$ and agent $a^2$ simultaneously attempt to move into each other’s positions, resulting in an edge conflict. Since $a^0$ holds higher priority, it occupies the desired vertex while forcing $a^2$ to adjust downward, thereby resolving the conflict locally. However, as illustrated in Fig.~\ref{fig:real-robot}(d), a livelock situation occurs when agent $a^0$ and agent $a^1$ oscillate at each other’s goal positions due to priority-based planning, which Extended-PIBT alone cannot resolve. At this point, CTS-PLL detects the livelock and invokes the LaCAM-based local re-planning module. As shown in Fig.~\ref{fig:real-robot}(e), the two agents are guided to a nearby region with connectivity degree three, where they successfully exchange positions. This recovery enables all agents to eventually complete their assigned tasks and reach their respective goals, as seen in Fig.~\ref{fig:real-robot}(f). These results demonstrate that CTS-PLL can effectively detect and resolve livelock situations, validating its robustness in real-world experiments.

\section{Conclusions}
\label{Sec:con}

This paper presented CTS-PLL, an enhanced framework for solving the CTS-MAPF problem by integrating configuration-based search with mechanisms for robustness and solution refinement. Through the combination of lock detection with LaCAM-based local re-planning, the algorithm effectively overcomes failure cases that limit CTS-PIBT. The inclusion of an anytime LNS refinement further improves path quality, particularly in dense and congested settings. Experimental results across diverse benchmarks confirm that CTS-PLL consistently outperforms prior approaches in both feasibility and efficiency. In the sparse setting, CTS-PLL achieves 100\% success, far surpassing CBSS and CTS-CBS, and improves solution quality by an average of 10.4\% over CTS-PIBT. In the dense scenario, two sets of ablation experiments effectively demonstrated the feasibility and efficiency improvement of CTS-PLL compared to CTS-PIBT. Physical robot tests further demonstrated the method’s applicability under real-world conditions. Future work will explore scaling the framework to larger agent-task systems, incorporating learning-based heuristics to accelerate decision-making, and extending the approach to dynamic or uncertain environments.

\addtolength{\textheight}{0cm}   







\bibliography{IEEEabrv, mybibfile}

\end{document}